%% file: main.tex
\documentclass{article}

\usepackage{conference,times}

\iclrfinalcopy

\usepackage{booktabs}
\usepackage{graphicx}
\usepackage{amsmath}
\usepackage{amssymb}
\usepackage{physics}
\usepackage{nicefrac}

\usepackage{multirow}

\usepackage{color}

\usepackage{subcaption}

\usepackage[normalem]{ulem}

\usepackage{hyperref}
\hypersetup{
    colorlinks=true,
    linkcolor=blue,
    citecolor=blue,
    urlcolor=blue
}
\usepackage[capitalize]{cleveref}
\crefname{algorithm}{Algorithm}{Algorithms}
\usepackage{url}

\title{
Adaptive Layer-Wise Transformations \\for Post-Training Quantization of \\ Large Language Models}

\author{
  \textbf{Cuong Pham}\textsuperscript{1} \quad
  \textbf{Hoang Anh Dung}\textsuperscript{1} \quad
  \textbf{Cuong C. Nguyen}\textsuperscript{2} \quad
  \textbf{Trung Le}\textsuperscript{1} \\
  \hfill
  \textbf{Gustavo Carneiro}\textsuperscript{2} \quad
  \textbf{Jianfei Cai}\textsuperscript{1} \quad
  \textbf{Thanh-Toan Do}\textsuperscript{1}
  \hfill\mbox{} \\[0.5em]
  \textsuperscript{1}Department of Data Science and AI, Monash University, Australia \\
  \textsuperscript{2}Centre for Vision, Speech and Signal Processing, University of Surrey, UK \\
}

\begin{document}

\maketitle

\input{tex/abstract}

\input{tex/introduction}

\input{tex/relatedworks}

\input{tex/method}

\input{tex/experiments}
\input{tex/conclusion}

\clearpage
\bibliography{reference}
\bibliographystyle{conference}

\end{document}

%% file: tex/abstract.tex
\begin{abstract}
    Large language models require significant computational resources for deployment, making quantization essential for practical applications. 
    However, the main obstacle to effective quantization lies in systematic outliers in activations and weights, which cause substantial LLM performance degradation, especially at low-bit settings.
    While existing transformation-based methods like affine and rotation transformations successfully mitigate outliers, they apply the homogeneous transformation setting, i.e., using the same transformation types across all layers, ignoring the heterogeneous distribution characteristics within LLMs. In this paper, we propose an adaptive transformation selection framework that systematically determines optimal transformations on a per-layer basis. 
    To this end, we first formulate transformation selection as a differentiable optimization problem to achieve the accurate transformation type for each layer.  
    However, searching for optimal layer-wise transformations for every model is computationally expensive.
    To this end, we establish the connection between weight distribution kurtosis and accurate transformation type.  Specifically, we propose an outlier-guided layer selection method using robust $z$-score normalization that achieves comparable performance to differentiable search with significantly reduced overhead. Comprehensive experiments on LLaMA family models demonstrate that our adaptive approach consistently outperforms the widely-used fixed transformation settings. 
    For example, our method achieves an improvement of up to 4.58 perplexity points and a 2.11\% gain in average six-task zero-shot accuracy under aggressive W3A3KV2 quantization settings for the LLaMA-3-8B model compared to the current best existing method, FlatQuant, demonstrating the necessity of heterogeneous transformation selection for optimal LLM quantization.
    \end{abstract}

%% file: tex/introduction.tex
\section{Introduction}

Large language models (LLMs)~\citep{Wei_2022,touvron2023llama,zhang2022opt} have gained significant attention due to their remarkable performance in handling complex natural language tasks~\citep{Hendrycks_2020}, such as language generation, translation, question answering, and text summarization.
However, their billion-parameter scale requires significant computational resources for inference and deployment~\citep{frantar2022gptq,lin2023awq}. 
Quantization, especially post-training quantization (PTQ), has emerged as the dominant compression technique, significantly reducing both memory footprint and computational requirements while preserving model capabilities~\citep{Dettmers,xiao2023smoothquant}.

The primary obstacle to quantizing LLMs effectively is the presence of outliers~\citep{an2025systematic} in activations and weight parameters. For example, \citet{wei2023outlier} point out that activation values that exceed the mean by 10-100 standard deviations appear consistently at specific channels across different inputs.
This causes significant quantization degradation, especially when quantized to extremely low-bit settings. Several methods have been proposed to address this challenge. In \citep{quik,cui2024cherry,kim2023squeezellm}, they propose mixed precision, which assigns some of the channels to higher precision and less sensitive channels to lower precision to balance accuracy and efficiency. Another approach is to apply transformations to the weights and activations to mitigate the outlier problem.

SmoothQuant~\citep{xiao2023smoothquant} and OmniQuant~\citep{shao2023omniquant} apply per-channel scaling to balance activation and weight magnitudes. However, these methods cannot effectively quantize the LLMs to 4-bit weights and activations. Affine transformation~\citep{AffineQuant,sun2024flatquant} and rotation transformation~\citep{ashkboos2024quarot,liu2025spinquant,hu2025ostquant} are proposed to mitigate the outlier problem effectively and successfully quantize the LLMs to the 4-bit setting.
Rotation transformation-based approaches~\citep{ashkboos2024quarot,liu2025spinquant,hu2025ostquant} apply orthogonal matrices to spread outliers across dimensions to mitigate the outlier problem. Meanwhile, affine transformation-based approaches~\citep{AffineQuant,sun2024flatquant} apply learnable matrices to flatten the distribution to mitigate the outlier problem.

While affine transformations theoretically offer greater flexibility than rotation transformations for handling outliers, the original AffineQuant~\citep{AffineQuant} approach has practical limitations. It learns a full transformation matrix that can only be applied to output projection layers for weight-activation quantization, where the matrix merges with preceding linear layers to avoid overhead. Other layers must use per-channel scaling, limiting the method's broader applicability across model architectures.
FlatQuant~\citep{sun2024flatquant} uses the Kronecker product to decompose the transformation into two smaller matrices, significantly reducing the number of learnable parameters and achieving the best performance among the existing transformation-based approaches for PTQ of LLMs.
Even though these approaches successfully quantize LLMs to 4 bits with slight performance degradation, they apply the same type of transformation across all layers, ignoring the distribution characteristics of each layer within LLMs.

Our key insight is that different layers exhibit fundamentally different statistical properties that determine their optimal transformation. Each layer has its own distribution characteristics; therefore, the optimal transformation type could be different for different layers.
Specifically, we propose an approach using differentiable search to find the optimal transformation type for each layer.
However, this approach is computationally expensive and not practical for large-scale models. To that end, we propose using outliers as a critical statistical measure for transformation selection. Specifically, we leverage kurtosis~\citep{decarlo1997meaning} to analyze weight outliers and scrutinize the relationship between kurtosis and the optimal transformation types. We observe that the relation between weight distribution kurtosis and the preferred transformation is layer-type dependent. For FFN layers, layers with higher kurtosis, whose concentrated outliers benefit from rotation's redistribution capability, tend to favor rotation transformations, while most low-kurtosis FFN layers favor efficient affine transformations. For attention layers, the trend is reversed. Layers with higher kurtosis are more often assigned affine transformations, while most low-kurtosis attention layers are assigned rotation transformations.
Based on this insight, we propose an outlier-guided transformation selection approach that leverages the kurtosis-based insights to efficiently identify the best transformation type for each layer.

In summary, the main contributions of this paper are as follows.
\begin{itemize}
    \item Unlike existing transformation-based approaches for handling outliers in LLM quantization, which typically apply the same transformation types across all layers, we introduce the novel idea of using layer-specific transformations and demonstrate the superior performance across multiple model architectures and quantization settings.

    \item We provide the connection between weight outliers and the optimal transformation type and propose an efficient heuristic outlier-guided transformation selection method that achieves comparable performance to differentiable search with significantly reduced computational overhead, making it practical for large-scale models where differentiable search becomes computationally prohibitive.
\end{itemize}

%% file: tex/relatedworks.tex
\section{Related Works and BACKGROUND}

\subsection{Related Works} 
\paragraph{Post-Training Quantization.}
Post-training quantization (PTQ) has become the dominant approach for compressing LLMs due to its efficiency and practicality compared to quantization-aware training. These methods apply quantization to pre-trained models using minimal calibration data, avoiding expensive retraining. {GPTQ}~\citep{frantar2022gptq} pioneered layer-wise quantization using second-order Hessian information for error compensation, achieving impressive compression rates. 
{AWQ}~\citep{lin2023awq} advances weight-only quantization by incorporating activation awareness, identifying salient weight channels based on activation magnitudes. Other notable PTQ methods include ZeroQuant~\citep{yao2022zeroquant}, which proposes fine-grained quantization schemes, and OWQ~\citep{lee2024owq}, which improves upon AWQ's activation-aware approach. Recent works like QuIP~\citep{chee2023quip} and QuIP\#~\citep{tseng2024quip} introduce vector quantization with incoherent processing, though at the cost of additional computational overhead.

\paragraph{Transformation-Based Approaches}
Transformation-based methods that modify model weights and activations are the most effective approaches for mitigating outliers in LLM quantization. SmoothQuant~\citep{xiao2023smoothquant} applies per-channel scaling to balance quantization difficulty between activations and weights.
Rotation transformations offer several key advantages for outlier mitigation. They preserve the Euclidean norm of weight vectors, maintaining the geometric structure of the original model. Additionally, they redistribute concentrated outliers across multiple dimensions through orthogonal matrix multiplication, effectively flattening the distribution without information loss. QuaRot~\citep{ashkboos2024quarot} is the first to apply rotation transformation in post-training quantization (PTQ) for LLMs and successfully quantizes LLMs to 4-bit weights and activations. Specifically, they adopt the Hadamard transform to redistribute outliers across dimensions. SpinQuant~\citep{liu2025spinquant} and OSTQuant~\citep{hu2025ostquant} further improve performance by learning optimal orthogonal matrices to better mitigate the outlier problem.
Meanwhile, affine transformations provide greater flexibility in reshaping distributions compared to rotation transformations. AffineQuant~\citep{AffineQuant} is the first to propose learnable affine transformations for outlier mitigation. FlatQuant~\citep{sun2024flatquant} further improves this approach by proposing a Kronecker-based affine transformation to reduce the number of learnable parameters and achieve significant improvements, becoming the best existing transformation-based approach.

\paragraph{Statistical Analysis in Quantization}

Understanding the statistical properties of weight and activation distributions is fundamental to effective quantization design. Prior work has shown that outliers in LLMs exhibit systematic patterns~\citep{wei2023outlier}, consistently emerging in specific channels across different inputs, with their magnitude and frequency strongly correlating with model scale and becoming increasingly pronounced beyond 6.7B parameters. However, existing research focuses primarily on outlier identification rather than understanding how statistical properties should guide transformation selection.
Our work addresses this gap by investigating how the statistical characteristics of outliers can be used to guide transformation selection. While kurtosis has been recognized as an indicator of distribution tailedness, the relationship between distribution characteristics and optimal transformation selection remains largely unexplored. We address this limitation by establishing kurtosis as the critical statistical measure for transformation selection.

\subsection{Background}
\label{subsec:problem_formulation_preliminary}

Given a pre-trained full-precision LLM, a layer of weights $\mathbf{W} \in \mathbb{R}^{m \times n}$ and input activations $\mathbf{X} \in \mathbb{R}^{b \times m}$, quantization maps these to lower-precision representations. The quantized output of the layer is given by:
\begin{equation}
\hat{\mathbf{Y}} = \mathcal{Q}_a(\mathbf{X}) \cdot \mathcal{Q}_w(\mathbf{W}),
\end{equation}
where $\mathcal{Q}_a$ and $\mathcal{Q}_w$ denote the quantization functions of the activations and weights, respectively. For a $k$-bit quantization, these functions can be expressed as follows:
\begin{equation}
\mathcal{Q}(\mathbf{Z}) = s \cdot \text{clip}\left(\text{round}\left(\frac{\mathbf{Z}}{s}\right), -2^{k-1}, 2^{k-1}-1\right),
\end{equation}
where $s$ is the quantization scaling factor.

Affine~\citep{AffineQuant,sun2024flatquant} and rotation~\citep{ashkboos2024quarot,liu2025spinquant,hu2025ostquant} are the two important transformation approaches to deal with the outliers for post-training quantization of LLMs. The affine transformation flattens distributions using a learnable matrix:
\begin{equation}
\hat{\mathbf{Y}}_{A} = \mathcal{Q}_a(\mathbf{X} \mathbf{A})\mathcal{Q}_w(\mathbf{A}^{-1} \mathbf{W}),
\end{equation}
where $\mathbf{A}$ is a learnable affine transformation matrix.

The rotation transformation uses orthogonal matrices for outlier redistribution:
\begin{equation}
\hat{\mathbf{Y}}_{R} = \mathcal{Q}_a(\mathbf{X} \mathbf{R}) \mathcal{Q}_w(\mathbf{R}^{\top} \mathbf{W}),
\end{equation}
where $\mathbf{R} \in \mathbb{R}^{m \times m}$ satisfies $\mathbf{R}^{\top}\mathbf{R} = \mathbf{I}$.

We demonstrate that different layers exhibit distinct statistical properties that make them more suitable for different transformation types. This motivates our adaptive selection approach.

%% file: tex/method.tex
\section{Proposed Method}

\subsection{Problem and Motivation}

Existing transformation-based methods for LLM quantization apply homogeneous transformations (either rotation or affine) across all layers, ignoring the heterogeneous distribution characteristics within LLMs. Different layers exhibit distinct statistical properties that make them more suitable for different transformation types. Rotation transformations preserve Euclidean norm and redistribute outliers through orthogonal matrices, while affine transformations offer greater flexibility in reshaping distributions. Therefore, selecting the optimal transformation for each layer could significantly improve quantization performance.

We present a novel transformation selection framework for post-training quantization (PTQ) of large language models (LLMs). We first present the motivation for adaptive transformation selection, then introduce a layer-wise transformation selection method using differentiable search. While effective, this approach is computationally expensive and impractical for large-scale applications. To address this limitation, we analyze the differentiable search results to identify correlations between weight distribution characteristics and selected transformation types. These insights enable us to develop a more efficient heuristic approach that not only maintains effectiveness but is also practical for large-scale models. Our approach combines both insights from statistical analysis and practical techniques to achieve state-of-the-art quantization performance.

\begin{table}[!h]
\centering
\caption{Comparison of the performance of adaptive transformation selection using LLaMA-2-7B for W3A3KV3 quantization setting. We report the mean and standard deviation of the performance over 20 random sets of the transformation selection as well as the best result from 20 random trials.}
\label{tab:motivation}
\resizebox{0.65\textwidth}{!}{
\begin{tabular}{l | c c | c}
\toprule
\textbf{Configuration} & \textbf{WikiText-2 ($\downarrow$)} & \textbf{C4 ($\downarrow$)} & \textbf{Zero-shot Avg ($\uparrow$)}  \\
\midrule
FP16 & 5.47 & 7.26 & 69.79  \\
\midrule
Fixed Affine Transformation & 7.54 & 9.76 & 59.89  \\
Fixed Rotation Transformation & 7.99 & 10.90 & 58.91  \\
Random Transformation Selection & 7.61 $\pm$ 0.36 & 9.96 $\pm$ 0.59 & 59.64 $\pm$ 0.75  \\
\midrule
\textbf{Best result} & \textbf{7.26} & \textbf{9.42} & \textbf{61.15}  \\
\bottomrule
\end{tabular}}
\end{table}

\paragraph{Preliminary analysis on the impact of the transformation selection.}
We conduct a preliminary analysis to study the effect of transformation matrices in post-training quantization (PTQ) for LLMs. We use the released code from FlatQuant~\citep{sun2024flatquant} to quantize LLaMA-2-7B with the affine transformation. To study the effect of the transformation matrices, we randomly assign 50\% of layers of attention and feed-forward layers to use the affine transformation and the remaining 50\% to use the rotation transformation. As shown in Table~\ref{tab:motivation}, the best result among 20 runs improves over the setting using all affine transformations by 0.28 and 0.34 perplexity scores on WikiText-2~\citep{merity2016pointer} and C4~\citep{raffel2020exploring}, respectively. Moreover, the performance on average accuracy across six zero-shot tasks significantly improves by 1.26 points. This implies that performance can be significantly improved by carefully selecting the transformation for each layer when performing post-training quantization for LLMs.

\subsection{Differentiable Search for Accurate Transform Selection}
{To preliminarily assess the potential performance of layer-wise adaptive transformation,} we formulate the transformation selection as a differentiable 
search problem. For each layer $l$, the parameters $\boldsymbol{\alpha}^{(l)} = [\alpha_A^{(l)}, \alpha_R^{(l)}]^{\top}$  control the mixture between affine and rotation transforms.

The quantized output for layer $l$ is:
\begin{equation}
\hat{\mathbf{Y}}^{(l)} = \pi_A^{(l)} \cdot \hat{\mathbf{Y}}_A^{(l)} + \pi_R^{(l)} \cdot \hat{\mathbf{Y}}_R^{(l)},
\end{equation}
where $\pi_t^{(l)} = \exp(\alpha_t^{(l)}) / \sum_{t'} \exp(\alpha_{t'}^{(l)})$ are softmax weights ensuring $\pi_A^{(l)} + \pi_R^{(l)} = 1$.

We optimize a loss function that combines reconstruction error and entropy regularization:
\begin{equation}
\mathcal{L} = \sum_{l} \left[ \mathcal{L}_{\text{recon}}^{(l)} + \lambda_{\text{entropy}} \mathcal{H}(\boldsymbol{\pi}^{(l)}) \right],
\end{equation}
where $\mathcal{L}_{\text{recon}}^{(l)} = \norm{\mathbf{Y}^{(l)} - \hat{\mathbf{Y}}^{(l)}}_F^2$ measures the reconstruction error between the full-precision layer output $\mathbf{Y}^{(l)}$ and the quantized output $\hat{\mathbf{Y}}^{(l)}$, $\mathcal{H}(\boldsymbol{\pi}^{(l)}) = -\sum_t \pi_t^{(l)} \log \pi_t^{(l)}$ is the entropy regularization that encourages the softmax weights to converge toward binary values (zero or one){, and \(\lambda_{\text{entropy}}\) is a hyper-parameter weighting the contribution of the entropy}. The selection variables for all layers are optimized jointly under this summed objective.

After convergence, we discretize the selection:
\begin{equation}
\mathbf{T}^{(l)} = \begin{cases}
\mathbf{A}^{(l)} & \text{if } \operatorname{argmax}_t \pi_t^{(l)} = A \\
\mathbf{R}^{(l)} & \text{if } \operatorname{argmax}_t \pi_t^{(l)} = R.
\end{cases}
\end{equation}


While differentiable search can achieve accurate transformation selection, its computational overhead makes it impractical for large-scale models. Therefore, we develop an efficient statistical framework to guide transformation selection based on layer-specific distribution properties.

\subsection{Correlation Between Kurtosis and Transformation Selection}
\label{subsec:layer_analysis}

While differentiable search yields accurate per-layer transform choices, it is computationally expensive for large models and potentially unnecessary. To that end, we propose an outlier-guided layer-wise transformation selection that leverages the \emph{kurtosis}, a statistical measure of a distribution's ``tailedness'', as an indicator to efficiently choose the best transformation for each layer.

We compute kurtosis as a metric to characterize weight distribution properties. For layer $l$ with weights $\mathbf{W}^{(l)}$, the excess kurtosis is:
\begin{equation}
\label{eq:kurtosis_score}
\kappa^{(l)} = \frac{\mathbb{E}[(\operatorname{vec}(\mathbf{W}^{(l)}) - \mu^{(l)})^4]}{(\sigma^{(l)})^4} - 3,
\end{equation}
where $\mu^{(l)}$ and $\sigma^{(l)}$ are the mean and standard deviation of the vectorized weights {and \(\operatorname{vec}(.)\) is the function flattening a matrix into a vector}.

A high value of the kurtosis (i.e., leptokurtic distributions) indicates heavy tails and the presence of outliers, while a low value of the kurtosis (i.e., platykurtic distributions) suggests a more uniform distribution~\citep{decarlo1997meaning}. 
We adopt kurtosis because it is tail-sensitive, directly reflects outliers, and correlates with the transformation choices found via differentiable search. This makes kurtosis a practical diagnostic for anticipating which layers benefit most from rotation or affine transforms.

Since kurtosis is tail-sensitive, it measures how strongly the weight values of a layer are concentrated in a few extreme values. The two transformations address outliers through different mechanisms. The rotation transformation applies orthogonal matrices to spread outliers across dimensions while preserving the Euclidean norm, so its benefit is largest when a layer has heavy tails with concentrated outliers. The affine transformation reshapes the entire distribution using a learnable matrix, so its greater flexibility is most useful when the distribution is already relatively flat and there are few concentrated outliers to spread.


\begin{figure}[t]
    \centering
    \begin{subfigure}{0.48\textwidth}
        \centering
        \includegraphics[width=\textwidth]{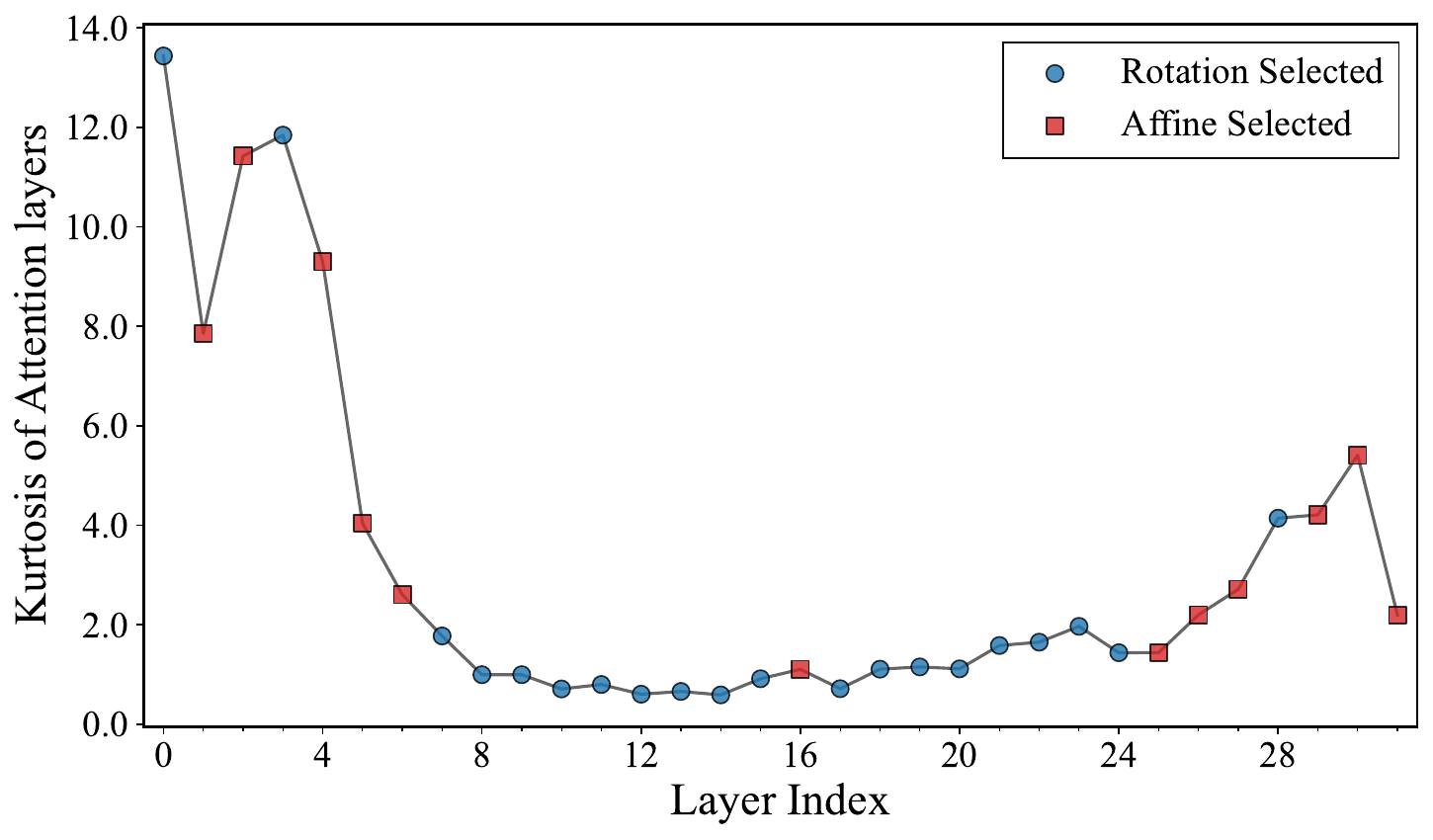}
        \caption{Kurtosis of Attention layers in LLaMA-2-7B}
        \label{fig:attention_outliers_ds}
    \end{subfigure}
    \hfill
    \begin{subfigure}{0.48\textwidth}
        \centering
        \includegraphics[width=\textwidth]{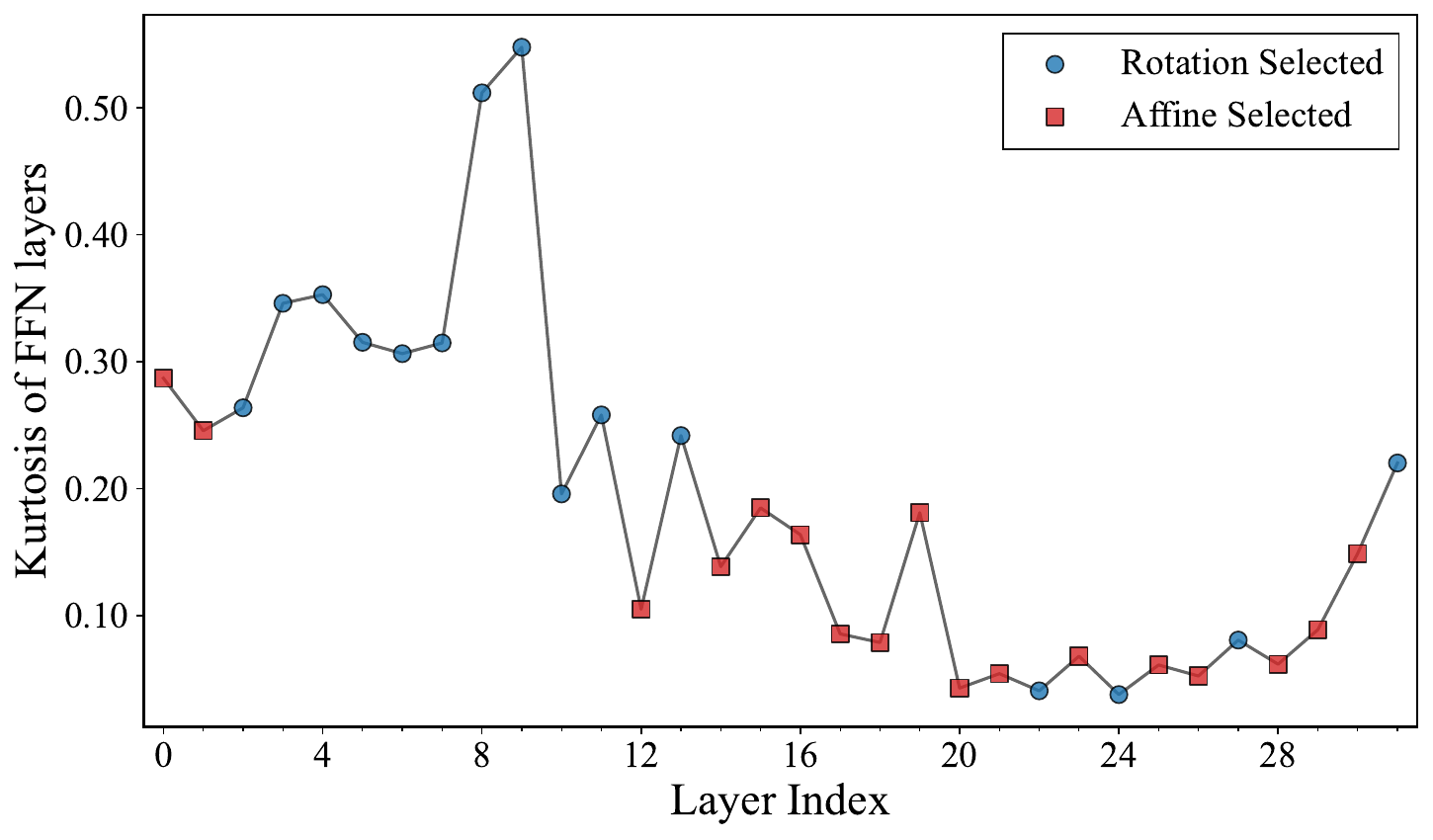}
        \caption{Kurtosis of FFN layers in LLaMA-2-7B}
        \label{fig:mlp_outliers_ds_sub}
    \end{subfigure}
    \caption{Kurtosis scores across layers for attention and FFN components with corresponding selected transformations using differentiable search. Blue circles indicate layers where rotation transformation was selected, while red squares indicate layers where affine transformation was selected based on differentiable search.}
    \label{fig:outliers_comparison}
    \end{figure}
To quantify the relationship between outliers and transformation selection, we compute kurtosis scores for each layer type. For attention layers, we calculate the sum of the kurtosis scores of Query (Q), Key (K), and Value (V) layers, while for feed-forward (FFN) layers, we compute the kurtosis score of the Gate/Up projection layer using Eq.~\ref{eq:kurtosis_score}. These kurtosis scores capture the concentration of extreme values in each layer type and provide a quantitative basis for understanding the correlation between statistical properties and transformation selection.

Specifically, Figure~\ref{fig:outliers_comparison} highlights a correlation between the kurtosis scores and the selected transformation. For the attention layers, layers with higher kurtosis scores are more likely to be selected for the affine transformation, while most layers with kurtosis scores less than $2.0$ are selected for the rotation transformation. 
Conversely, for the FFN layers, layers with higher kurtosis scores are more likely to be selected for the rotation transformation, while most layers with kurtosis scores less than $0.2$ are selected for the affine transformation. Additionally, for each type of layer, the rotation or affine transformations are selected at both high and low kurtosis scores, indicating that kurtosis provides guidance rather than strict thresholds for transformation selection.


To bridge the gap between statistical understanding and practical implementation, we propose an outlier-guided layer-wise transformation selection that leverages the kurtosis metric as an indicator to efficiently identify the best transformation for each layer. 

\subsection{Outlier-Guided Transformation Selection}

Given the model comprises $n$ sequential attention (or feed-forward) layers indexed $i\in\{0,\dots,n-1\}$, we treat the kurtosis value as an outlier indicator and set $o_i = |\kappa^{(i)}|$, which is the absolute value of the kurtosis score of the layer,
as the layer's outlier score. Our goal is to {leverage \(\{o_{i}\}\) to} select $L$ $(1\le L\le n)$ layer indices as the rotation transformation selection and the remaining $n-L$ layer indices as the affine transformation selection.

\paragraph{Robust normalization.}
To achieve scale- and skew-robustness, we convert {the absolute kurtosis scores} $\{o_i\}$ to the robust $z$-scores \citep{iglewicz1993volume} via the median and the median absolute deviation (MAD):
\begin{equation}
\tilde{o}_i \;=\; \frac{o_i - \mathrm{median}(o)}{1.4826\,\mathrm{MAD}(o) + \varepsilon},
\qquad
\mathrm{MAD}(o) \;=\; \mathrm{median} \abs{ o - \mathrm{median}(o) },
\label{eq:robust_z}
\end{equation}
with a small value $\varepsilon>0$ (e.g., $10^{-12}$) for numerical stability. The factor $1.4826$ scales MAD to be comparable to a standard deviation under normality.

Empirically, rotation transformations are concentrated in both high and low kurtosis scores, with a prevalence at higher kurtosis scores for FFN layers and a prevalence in lower kurtosis scores for attention layers.
We therefore allocate a fraction $\beta$ of the $L$ layers as the rotation transformation to the higher part of kurtosis scores:
\begin{equation}
K_{\mathrm{high}}=\bigl\lfloor \beta L \bigr\rceil,\qquad
K_{\mathrm{low}} = L - K_{\mathrm{high}},
\label{eq:budget_split}
\end{equation}
where $\lfloor . \rceil$ denotes the rounding-to-nearest function.

Optionally, $\beta$ can be set by the positive-vs-absolute $z$-mass:
For rotation transformation selection at attention layers, we set $\beta_{\text{attn}}$ as:
\begin{equation}
\beta_{\text{attn}} \;=\; \mathrm{clip}\!\left(
\frac{\sum_{i:\,\tilde{o}_i>0}\tilde{o}_i}{\sum_{i}|\tilde{o}_i|},
\,0.1,\,0.3
\right).
\label{eq:alpha_attn}
\end{equation}

For rotation transformation selection at FFN layers, we set $\beta_{\text{ffn}}$ as:
\begin{equation}
\beta_{\text{ffn}} \;=\; \mathrm{clip}\!\left(
\frac{\sum_{i:\,\tilde{o}_i>0}\tilde{o}_i}{\sum_{i}|\tilde{o}_i|},
\,0.7,\,0.9
\right).
\label{eq:alpha_ffn}
\end{equation}

The ranges were selected from the LLaMA-2-7B W4A4KV2 ablations in \cref{tab:param_beta_attn,tab:param_beta_ffn}. In the experiments, we use the fixed values $\beta_{\text{attn}}=0.1$ and $\beta_{\text{ffn}}=0.9$ for all evaluated LLaMA models and quantization settings. Their transfer to other model families has not been evaluated and may require retuning.

We define upper and lower thresholds using order statistics of the outlier scores $\{\tilde{o}_i\}_{i=0}^{n-1}$. For the upper tail, we select the $(n-K_{\mathrm{high}})$-th largest value as the threshold:
\begin{equation}
\tau_{\mathrm{high}} = \begin{cases}
\text{$(n-K_{\mathrm{high}})$-th largest value in } \{\tilde{o}_i\}_{i=0}^{n-1} & \text{if } K_{\mathrm{high}} > 0 \\
+\infty & \text{if } K_{\mathrm{high}} = 0
\end{cases}
\label{eq:tau_high}
\end{equation}
Similarly, for the lower tail, we select the $(K_{\mathrm{low}})$-th smallest value:
\begin{equation}
\tau_{\mathrm{low}} = \begin{cases}
\text{$(K_{\mathrm{low}})$-th smallest value in } \{\tilde{o}_i\}_{i=0}^{n-1} & \text{if } K_{\mathrm{low}} > 0 \\
-\infty & \text{if } K_{\mathrm{low}} = 0
\end{cases}
\label{eq:tau_low}
\end{equation}

The candidate index set for the selection of rotation transformation is the union of the two tails:
\begin{equation}
\mathcal{C}
\;=\;
\bigl\{\,i:\tilde{o}_i \ge \tau_{\mathrm{high}}\,\bigr\}
\;\cup\;
\bigl\{\,i:\tilde{o}_i \le \tau_{\mathrm{low}}\,\bigr\}.
\label{eq:candidate_set}
\end{equation}

The heuristic integrates directly with existing quantization approaches, requiring only statistical classification to determine transform types and is then applicable to large-scale models 
compared to the differentiable search.

%% file: tex/experiments.tex
\section{Experiments}

\subsection{Experimental Setup}

\paragraph{Models and Datasets.} We evaluate on the LLaMA family: LLaMA-2 (7B, 13B)~\citep{touvron2023llama} and LLaMA-3 (8B), covering diverse model scales. Following previous works~\citep{shao2023omniquant, ashkboos2024quarot}, we report perplexity (PPL) on WikiText-2~\citep{merity2016pointer} and C4~\citep{raffel2020exploring} test sets for language modeling evaluation. For downstream task evaluation, we assess model performance on six zero-shot tasks using the \texttt{lm-evaluation-harness} framework, including ARC-Easy and ARC-Challenge~\citep{arc}, HellaSwag~\citep{hellaswag}, LAMBADA~\citep{paperno2016lambada}, PIQA~\citep{piqa}, and WinoGrande~\citep{sakaguchi2021winogrande}. We report the average accuracy across six zero-shot tasks as the primary evaluation metric for downstream performance. For calibration, we randomly sample 128 sequences, each containing 2048 tokens, from the WikiText-2 dataset \citep{merity2016pointer}. 

\paragraph{Baselines.} We compare against state-of-the-art PTQ methods representing different transformation approaches: 
SmoothQuant~\citep{xiao2023smoothquant}, affine transformation method FlatQuant~\citep{sun2024flatquant}, and rotation-based transformation methods QuaRot~\citep{ashkboos2024quarot}, SpinQuant~\citep{liu2025spinquant}, and OSTQuant~\citep{hu2025ostquant}. These baselines represent the spectrum of current PTQ approaches, from traditional weight quantization to advanced transformation-based methods.

\paragraph{Implementation Details.} Our method is implemented in PyTorch with HuggingFace Transformers. For the rotation transformation, we initialize the rotation matrix with random orthogonal matrices and adopt RiemannAdam~\citep{becigneul2018riemannian} as the optimizer for the learnable rotation matrix. For the learnable affine transformation, we use the same implementation as FlatQuant~\citep{sun2024flatquant}. We use the AdamW optimizer with a learning rate of $5 \times 10^{-3}$. For differentiable search, we set the entropy regularization to $\lambda_{\text{entropy}}=0.01$. We employ symmetric per-channel weight quantization and per-token activation quantization with GPTQ weight quantizers following existing approaches~\citep{sun2024flatquant,hu2025ostquant}. Following~\citep{sun2024flatquant}, we also employ the combination of scaling transformation with the selected transformation and adopt learnable clipping thresholds~\citep{shao2023omniquant} for weights and activations to eliminate outliers. For all experiments, we empirically set $\beta_{\text{attn}} = 0.1$ and $\beta_{\text{ffn}} = 0.9$ for the outlier-guided transformation selection method. The parameter $L$ in Eq.~\ref{eq:budget_split} is set to $0.7 \times n$ for attention layers and set to $0.5 \times n$ for feed-forward layers, where $n$ is the number of model attention (or feed-forward) layers. We conduct an ablation study on these hyper-parameters in Section~\ref{sec:ablation_hyperparams}. The notation \emph{W4A4KV4} denotes quantization with 4-bit weights and 4-bit activations, where \emph{KV4} indicates 4-bit quantization for the key and value projection layers in attention modules. For the positioning of the adaptive transformation, we selectively apply the proposed affine transformation to QKV layers in attention modules and up-gate layers in FFN modules. For all other linear layers, we follow the FlatQuant approach~\citep{sun2024flatquant}. This selective design choice preserves runtime speedup while achieving significant performance improvements over FlatQuant.

\begin{table}[th]
    \centering
    \small
    \setlength{\tabcolsep}{4pt}
    \caption{Perplexity scores ($\downarrow$) on WikiText-2 and C4 datasets for various quantization settings on LLaMA models.}
    \label{tab:adaptive_comparison}
    \resizebox{0.55\textwidth}{!}{
    \begin{tabular}{llcccccc}
    \toprule
    \multirow{2}{*}{\bfseries Setting} & \multirow{2}{*}{\bfseries Method} & \multicolumn{3}{c}{\bfseries WikiText-2 ($\downarrow$)} & \multicolumn{3}{c}{\bfseries C4 ($\downarrow$)} \\
    \cmidrule(lr){3-5} \cmidrule(lr){6-8}
    & & \bfseries 2-7B & \bfseries 2-13B & \bfseries 3-8B & \bfseries 2-7B & \bfseries 2-13B & \bfseries 3-8B \\
    \midrule
     & FP16 & 5.47 & 4.88 & 6.14 & 7.26 & 6.73 & 9.45 \\
    \midrule
    \multirow{5}{*}{W4A4KV4} & QuaRot & 6.10 & 5.40 & 8.16 & 8.32 & 7.54 & 13.38 \\
    & SpinQuant & 5.96 & 5.24 & 7.39 & 8.28 & 7.48 & 12.19 \\
    & OSTQuant & 5.91 & 5.25 & 7.29 & -- & -- & -- \\
    & FlatQuant & 5.78 & 5.11 & 6.90 & 7.86 & 7.11 & 11.21 \\
    & \textbf{Ours} & \textbf{5.61} & \textbf{4.97} & \textbf{6.89} & \textbf{7.58} & \textbf{6.74} & \textbf{10.08} \\
    \midrule
    \multirow{3}{*}{W3A3KV3} & OSTQuant & 8.32 & 6.86 & 13.59 & 11.42 & 9.45 & 21.06 \\
    & FlatQuant & 7.54 & 6.14 & 10.62 & 9.76 & 8.15 & 16.20 \\
    & \textbf{Ours} & \textbf{7.22} & \textbf{6.02} & \textbf{10.07} & \textbf{9.43} & \textbf{8.08} & \textbf{15.40} \\
    \midrule
    \multirow{3}{*}{W4A4KV2} & OSTQuant & 8.32 & 6.65 & 12.88 & 11.14 & 9.03 & 19.84 \\
    & FlatQuant & 7.51 & 5.98 & 8.73 & 9.87 & 8.11 & 13.86 \\
    & \textbf{Ours} & \textbf{6.98} & \textbf{5.79} & \textbf{8.25} & \textbf{9.04} & \textbf{7.66} & \textbf{12.59} \\
    \midrule
    \multirow{3}{*}{W3A3KV2} & OSTQuant & 12.28 & 8.97 & 26.48 & 17.36 & 12.65 & 43.62 \\
    & FlatQuant & 11.51 & 7.86 & 16.38 & 15.89 & 10.91 & 27.29 \\
    & \textbf{Ours} & \textbf{9.83} & \textbf{7.15} & \textbf{13.74} & \textbf{13.35} & \textbf{9.76} & \textbf{22.71} \\
    \bottomrule
    \end{tabular}}
\end{table}

\begin{table}[!h]
\centering
\caption{Zero-shot QA task results of 4-bit and 3-bit weight \& activation quantized LLaMA models.}
\label{tab:zero-shot-qa-combined}
\resizebox{0.95\linewidth}{!}
{
\begin{tabular}{c|l|ccccccc}
\toprule
\multicolumn{1}{c|}{\textbf{Model}} & \multicolumn{1}{l|}{\textbf{Method}} & \textbf{ARC-C ($\uparrow$)} & \textbf{ARC-E ($\uparrow$)}   & \textbf{HellaSwag ($\uparrow$)}   & \textbf{LAMBADA ($\uparrow$)}   & \textbf{PIQA ($\uparrow$)} & \textbf{WinoGrande ($\uparrow$)} &  \textbf{Avg ($\uparrow$)}     \\ 
\midrule

\multicolumn{9}{c}{\textit{W4A4KV4 Settings}}  \\
\midrule
\multirow{5}{*}{\textbf{LLaMA-2-7B}}  
& FP16 & 46.16 & 74.54 & 75.98 & 73.92 & 79.05 & 69.06 & 69.79 \\
\cmidrule{2-9}
& QuaRot & 42.32 & 68.35 & 72.53 & 65.40 & 76.33 & 65.11 & 65.01 \\
& SpinQuant & 41.72 & 69.28 & 72.90 & 71.28 & 76.17 & 66.06 & 66.23 \\
& FlatQuant & 43.00 & 71.21 & 73.31 & \textbf{72.06} & 77.53 & 67.72 & 67.47 \\
& Ours & \textbf{43.94} & \textbf{72.31} & \textbf{73.66} & {71.38} & \textbf{77.64} & \textbf{68.35} & \textbf{67.88} \\
\midrule

\multirow{5}{*}{\textbf{LLaMA-2-13B}}  
& FP16 & 49.15 & 77.44 & 79.39 & 76.73 & 80.47 & 72.14 & 72.55 \\
\cmidrule{2-9}
& QuaRot & 45.48 & 73.27 & 76.03 & 69.01 & 79.05 & 70.64 & 68.91 \\
& SpinQuant & 49.15 & 77.19 & 76.86 & 73.86 & 78.67 & 69.85 & 70.93 \\  
& FlatQuant & 48.38 & \textbf{76.94} & 77.88 & \textbf{76.40} & 79.65 & 70.56 & 71.64 \\
& Ours & \textbf{49.91} & {75.93} & \textbf{77.92} & {76.03} & \textbf{79.71} & \textbf{70.64} & \textbf{71.69} \\
\midrule

\multirow{5}{*}{\textbf{LLaMA-3-8B}}  
& FP16 & 53.50 & 77.57 & 79.12 & 75.51 & 80.74 & 72.93 & 73.23 \\
\cmidrule{2-9}
& QuaRot & 45.73 & 70.83 & 72.97 & 62.70 & 75.35 & 67.17 & 65.79 \\
& SpinQuant & 47.27 & 74.20 & 74.55 & 70.29 & 77.37 & 68.51 & 68.70 \\
& FlatQuant & 50.51 & 75.88 & 76.49 & \textbf{73.20} & \textbf{79.00} & \textbf{72.93} & 71.33 \\
& Ours & \textbf{51.45} & \textbf{78.16} & \textbf{76.50} & {72.50} & {78.89} & {72.38} & \textbf{71.65} \\
\midrule
\multicolumn{9}{c}{\textit{W3A3KV3 Settings}}  \\
\midrule
\multirow{3}{*}{\textbf{LLaMA-2-7B}}  
& OSTQuant & 30.72 & 55.09 & 60.21 & 57.05 & 68.44 & 57.77 & 54.88 \\
& FlatQuant & 36.35 & 60.82 & 66.20 & 60.59 & 73.29 & 59.43 & 59.45 \\
& Ours & \textbf{36.52} & \textbf{63.17} & \textbf{66.42} & \textbf{62.91} & \textbf{74.43} & \textbf{63.61} & \textbf{61.18} \\
\midrule
\multirow{3}{*}{\textbf{LLaMA-2-13B}}  
& OSTQuant & 37.37 & 61.41 & 66.44 & 62.12 & 71.98 & 58.01 & 59.56 \\  
& FlatQuant &40.96 &  70.41 & 71.92 & 71.41 & \textbf{77.20} &  \textbf{68.90} & 66.80 \\
& Ours & \textbf{42.83} & \textbf{71.97} & \textbf{72.33} & \textbf{71.71} & {76.77} &{66.14} & \textbf{66.94} \\
\midrule
\multirow{3}{*}{\textbf{LLaMA-3-8B}}  
& OSTQuant & 31.83 & 52.15 & 56.72 & 45.64 & 66.49 & 55.09 & 51.32 \\
& FlatQuant & 37.46 & 63.09 & 64.86 & \textbf{55.54} & 71.17 & 61.40 & 59.01 \\
& Ours & \textbf{38.05} & \textbf{65.19} & \textbf{65.71} & {52.73} & \textbf{73.29} & \textbf{62.98} & \textbf{59.66} \\

\midrule
\multicolumn{9}{c}{\textit{W3A3KV2 Settings}}  \\
\midrule
\multirow{3}{*}{\textbf{LLaMA-2-7B}}  
& OSTQuant & 26.54 & 46.46 & 48.11 & 34.45 & 64.58 & 52.64 & 45.46 \\
& FlatQuant & 30.89  & 50.63 & 56.79 & 32.95 & 67.08 & 52.72 & 48.51 \\
& Ours & \textbf{32.85} & \textbf{57.37} & \textbf{59.89} & \textbf{47.45} & \textbf{71.98} & \textbf{54.46} & \textbf{54.00} \\
\midrule
\multirow{3}{*}{\textbf{LLaMA-2-13B}}  
& OSTQuant & 31.66 & 52.69 & 58.43 & 45.41 & 67.36 & 53.28 & 51.47 \\  
& FlatQuant & 36.18 &  64.18 & 65.80 & 52.57 & \textbf{73.72} & 59.12& 58.60 \\
& Ours & \textbf{37.71} & \textbf{65.82} & \textbf{66.65} & \textbf{58.49} & 73.23 & \textbf{60.14} & \textbf{60.34} \\
\midrule
\multirow{3}{*}{\textbf{LLaMA-3-8B}}  
& OSTQuant & 24.74 & 38.89 & 40.18 & 18.07 & 56.75 & 49.01 & 37.94 \\
& FlatQuant & 33.02 & 56.23 & 55.48 & 25.79 & 69.31 & 56.75 & 49.43 \\
 & Ours & \textbf{34.64} & \textbf{56.90} & \textbf{57.86} & \textbf{32.25} & \textbf{69.75} & \textbf{57.85} & \textbf{51.54} \\
\bottomrule
\end{tabular}}
\end{table}

\subsection{Experimental Results}

\paragraph{Evaluation on generation datasets with perplexity.} We compare our proposed adaptive transformation selection method against state-of-the-art PTQ approaches, including QuaRot~\citep{ashkboos2024quarot}, SpinQuant~\citep{liu2025spinquant}, FlatQuant~\citep{sun2024flatquant}, and OSTQuant~\citep{hu2025ostquant}. 
The results of these methods are reproduced using their official implementations and are cited from FlatQuant~\citep{sun2024flatquant}. Table~\ref{tab:adaptive_comparison} presents the comparative results of our proposed method and other state-of-the-art approaches when evaluating on WikiText-2 and C4 datasets.  Our results are obtained using the proposed efficient heuristic outlier-guided transformation selection approach. We conduct experiments across multiple quantization configurations: W4A4KV4, W3A3KV3, W4A4KV2, and W3A3KV2. In general, our proposed method outperforms the state-of-the-art methods across various model architectures and quantization settings. Compared to the best existing method, FlatQuant~\citep{sun2024flatquant}, our proposed method consistently outperforms FlatQuant in all bit-width configurations. The improvement is clearer in extreme quantization settings, with improvements over FlatQuant of 0.55 and 0.80 points for LLaMA-3-8B in the W3A3KV3 setting on WikiText-2 and C4 datasets, respectively. Under the extremely low-bit W3A3KV2 quantization setting, our proposed method demonstrates substantial gains over FlatQuant with improvements of 1.68 and 2.54 perplexity points for LLaMA-2-7B, and 2.64 and 4.58 points for LLaMA-3-8B on WikiText-2 and C4 datasets, respectively, which confirms the effectiveness of our adaptive transformation selection approach.

\paragraph{Evaluation on downstream tasks.}
Table~\ref{tab:zero-shot-qa-combined} presents the comparative results of our proposed method and other state-of-the-art approaches across different quantization settings. The table evaluates performance on six tasks: ARC-Challenge, ARC-Easy, HellaSwag, LAMBADA, PIQA, and WinoGrande, with results averaged across all tasks. Under the W3A3KV3 setting, our method demonstrates significant improvements over existing approaches.  For LLaMA-2-7B, our approach achieves an average accuracy of 61.18\%, representing a substantial 1.73\% improvement over FlatQuant and a 6.30\% improvement over OSTQuant. The improvement is clearer in extreme W3A3KV2 quantization settings, with improvements of 5.49\% and 8.54\% for LLaMA-2-7B in the W3A3KV2 setting compared to FlatQuant and OSTQuant, respectively. For LLaMA-2-13B, our method achieves 60.34\%, outperforming FlatQuant by 1.74\% and OSTQuant by 8.87\%.
These results highlight the effectiveness of our adaptive transformation selection method in preserving model performance on downstream tasks, especially under extreme quantization scenarios.

\subsection{Ablation Studies}

\paragraph{The comparison between the proposed heuristic and the differentiable search selection.} 
Table~\ref{tab:kurtosis_heuristic} shows that the proposed outlier-guided transformation selection achieves 87.5\% and 85.0\% agreement with learned selection for LLaMA-2-7B and LLaMA-2-13B, respectively. Importantly, the proposed heuristic significantly reduces the training time by approximately 3$\times$ compared to the learned selection while maintaining competitive performance.

\begin{table}[!h]
\centering
\caption{Comparison of the proposed heuristic and differentiable search selection on LLaMA-2-7B and LLaMA-2-13B with W3A3KV3 settings.}
\label{tab:kurtosis_heuristic}
\resizebox{0.95\textwidth}{!}{
\begin{tabular}{l|cc|cc|cc|cc|cc}
\hline
\multirow{2}{*}{\textbf{Model}} & \multicolumn{2}{c|}{\textbf{WikiText-2 PPL}} & \multicolumn{2}{c|}{\textbf{C4 PPL}} & \multicolumn{2}{c|}{\textbf{Zero-shot Avg}} & \multicolumn{2}{c|}{\textbf{Selection Agreement}} & \multicolumn{2}{c}{\textbf{Training Time (h)}} \\
\cmidrule{2-11}
 & \textbf{Learned} & \textbf{Heuristic} & \textbf{Learned} & \textbf{Heuristic} & \textbf{Learned} & \textbf{Heuristic} & \textbf{Layers} & \textbf{Percentage} & \textbf{Learned} & \textbf{Heuristic} \\
\hline
LLaMA-2-7B & 7.18 & 7.22 & 9.35 & 9.43 & 61.45 & 61.18 & 28/32 & 87.5\% & 11 hours & 4 hours \\
LLaMA-2-13B & 5.99 & 6.02 & 8.03 & 8.08 & 67.04 & 66.94 & 34/40 & 85.0\% & 22 hours & 7.5 hours \\
\hline
\end{tabular}}
\end{table}

\paragraph{End-to-end model speedup.}

\begin{table}[!h]
    \centering
    \caption{End-to-end model speedup compared to FP16 on LLaMA-2-7B with a prefill sequence length of 2048 followed by 256 decoded tokens, measured on A100 GPUs including all online transformation overhead at inference. We assign the rotation per layer following our adaptive selection approach, covering 22 attention layers and 16 FFN layers. The remaining layers keep FlatQuant's Kronecker transformations.}
    \label{tab:model_speedup_bsz}
    \begin{minipage}{0.48\textwidth}
        \centering
        \subcaption{Batch size 1}
        \label{tab:model_speedup_bs1}
        \resizebox{\textwidth}{!}{
        \begin{tabular}{c|ccc}
            \toprule
            \textbf{Method}  & \textbf{Prefill} & \textbf{Decoding} & \textbf{End-to-End} \\
            \midrule
            INT4            & 1.203$\times$ & 0.620$\times$ & 0.666$\times$ \\
            QuaRot          & 1.071$\times$ & 0.562$\times$ & 0.566$\times$ \\
            FlatQuant       & 1.176$\times$ & 0.427$\times$ & 0.483$\times$ \\
            Ours            & 1.080$\times$ & 0.432$\times$ & 0.432$\times$ \\
            \bottomrule
        \end{tabular}}
    \end{minipage}
    \hfill
    \begin{minipage}{0.48\textwidth}
        \centering
        \subcaption{Batch size 16}
        \label{tab:model_speedup_bs16}
        \resizebox{\textwidth}{!}{
        \begin{tabular}{c|ccc}
            \toprule
            \textbf{Method}  & \textbf{Prefill} & \textbf{Decoding} & \textbf{End-to-End} \\
            \midrule
            INT4            & 1.236$\times$ & 0.735$\times$ & 0.784$\times$ \\
            QuaRot          & 1.042$\times$ & 0.644$\times$ & 0.675$\times$ \\
            FlatQuant       & 1.211$\times$ & 0.566$\times$ & 0.581$\times$ \\
            Ours            & 1.107$\times$ & 0.536$\times$ & 0.536$\times$ \\
            \bottomrule
        \end{tabular}}
    \end{minipage}
\end{table}

Table~\ref{tab:model_speedup_bsz} presents the prefill, decoding, and end-to-end speedup results for LLaMA-2-7B at batch sizes 1 and 16. All experiments were conducted on A100 GPUs under the W4A4KV4 setting, and each reported number is the mean over 10 timing repetitions after GPU warm-up. All timing measurements include the online transformation overhead incurred during inference. As shown, for prefill, our method achieves up to 1.11$\times$ speedup at batch size 16, surpassing QuaRot's 1.04$\times$ and approaching FlatQuant's 1.21$\times$. For decoding, our method is on par with FlatQuant at batch size 1 and stays within 0.03$\times$ at batch size 16. Our method attains an end-to-end speedup of 0.43$\times$ at batch size 1 and 0.54$\times$ at batch size 16, within approximately 0.05$\times$ of FlatQuant in both cases. These results demonstrate that our adaptive transformation selection approach effectively balances quantization performance and speedup efficiency.

\subsection{Ablation Studies of Hyper-parameter Settings}
\label{sec:ablation_hyperparams}

We conduct ablation studies to analyze the impact of the hyper-parameters $\beta_{\text{attn}}$ and $\beta_{\text{ffn}}$ in Eq.~\ref{eq:alpha_attn} and Eq.~\ref{eq:alpha_ffn}, respectively, in our adaptive layer-wise transformation scheme for post-training quantization of large language models. These parameters control the percentage of layers that select rotation transformations in the upper tail of z-scores of the attention and feed-forward layers.

\begin{table}[h]
    \centering
      \caption{Ablation studies for the hyper-parameter $\beta_{\text{attn}}$ in Eq.~\ref{eq:alpha_attn}. Results are on the LLaMA-2-7B model using W4A4KV2 quantization.}
      \label{tab:param_beta_attn}
    \begin{tabular}{l|cccccc}
      \toprule
       $\beta_{\text{attn}}$ & 0.1 & 0.3 & 0.4 & 0.5 & 0.7 & 0.9 \\
       \midrule
        WikiText-2 PPL ($\downarrow$) & \textbf{7.05} & 7.07 & 7.09 & 7.11 & 7.13 & 7.22 \\
        C4 PPL ($\downarrow$) & \textbf{9.12} & 9.12 & 9.16 & 9.12 & 9.22 & 9.37 \\
        Six zero-shot avg ($\uparrow$) & \textbf{61.40} & 61.11 & 61.20 & 60.09 & 60.70 & 61.05 \\
      \bottomrule
    \end{tabular}
\end{table}

\paragraph{Ablation studies for the hyper-parameter $\beta_{\text{attn}}$ in Eq.~\ref{eq:alpha_attn}.} For $\beta_{\text{attn}}$, we vary the values from 0.1 to 0.9 and evaluate the model performance on WikiText-2 and C4 datasets using perplexity (PPL) scores, as well as the average zero-shot performance across six downstream tasks. As shown in Table~\ref{tab:param_beta_attn}, smaller values of $\beta_{\text{attn}}$ (0.1-0.3) generally yield better performance, with $\beta_{\text{attn}} = 0.1$ achieving the best results across all metrics. This suggests that in attention layers, most selected rotation transformations are from the lower tail of the z-scores distribution.

\begin{table}[h]
    \centering
      \caption{Ablation studies for the hyper-parameter $\beta_{\text{ffn}}$ in Eq.~\ref{eq:alpha_ffn}. Results are on the LLaMA-2-7B model using W4A4KV2 quantization.}
      \label{tab:param_beta_ffn}
    \begin{tabular}{l|cccccc}
      \toprule
       $\beta_{\text{ffn}}$ & 0.1 & 0.3 & 0.5 & 0.7 & 0.8 & 0.9 \\
       \midrule
        WikiText-2 PPL ($\downarrow$) & 7.30 & 7.26 & 7.25 & 7.21 & 7.23 & \textbf{7.08} \\
        C4 PPL ($\downarrow$) & 9.54 & 9.46 & 9.41 & 9.41 & 9.42 & \textbf{9.37} \\
        Six zero-shot avg ($\uparrow$) & 60.83 & 61.07 & 61.16 & 61.19 & 61.26 & \textbf{61.30} \\
      \bottomrule
    \end{tabular}
\end{table}

\paragraph{Ablation studies for the hyper-parameter $\beta_{\text{ffn}}$ in Eq.~\ref{eq:alpha_ffn}.} For $\beta_{\text{ffn}}$, we test values ranging from 0.1 to 0.9. As shown in Table~\ref{tab:param_beta_ffn}, larger values of $\beta_{\text{ffn}}$ (0.7-0.9) generally yield better performance. The results indicate that $\beta_{\text{ffn}} = 0.9$ provides the optimal balance, achieving the lowest perplexity scores and highest zero-shot performance. This finding suggests that the feed-forward layers benefit from selecting rotation transformations predominantly from the upper tail of the z-scores distribution.

\paragraph{Ablation studies for the hyper-parameter $L$ in Eq.~\ref{eq:budget_split}.}
For the number of layers selecting the rotation transformation parameters $L$, we vary the value of $L$ from $0.1\times n$ to $0.9\times n$, where $n$ is the total number of attention layers. Following the ablation studies of $\beta_{\text{attn}}$ and $\beta_{\text{ffn}}$, we evaluate the model performance on the LLaMA-2-7B model using W4A4KV2 quantization. We set the number of selected rotation transformations of the feed-forward layers to zero, so that the results only show the effect of the attention layers.
As shown in Table~\ref{tab:param_L_attn}, the perplexity scores on WikiText-2 and C4 decrease when more attention layers are assigned the rotation transformation. Meanwhile, the average zero-shot performance across six downstream tasks achieves the best result when $L$ is set to $0.7\times n$, and does not improve further at $0.9\times n$. This suggests that selecting rotation transformations for 70\% of the attention layers is sufficient, while the remaining layers can use the affine transformation. Since the average zero-shot performance is the primary evaluation metric for downstream tasks, we set $L$ to $0.7\times n$ for the attention layers. For the feed-forward layers, the rotation transformation introduces a higher inference cost than the Kronecker-decomposed affine transformation used in FlatQuant~\citep{sun2024flatquant}. We therefore use a smaller number of selected rotation transformations and set $L$ to $0.5\times n$ for the feed-forward layers in all the experiments.

\begin{table}[h]
    \centering
      \caption{Ablation studies for the number of selected rotation transformations $L$ for attention layers in Eq.~\ref{eq:budget_split}. Results are on the LLaMA-2-7B model using W4A4KV2 quantization, with the feed-forward rotation budget set to zero.}
      \label{tab:param_L_attn}
    \begin{tabular}{l|ccccc}
      \toprule
       $L$ & $0.1\times n$ & $0.3\times n$ & $0.5\times n$ & $0.7\times n$ & $0.9\times n$ \\
       \midrule
        WikiText-2 PPL ($\downarrow$) & 7.41 & 7.13 & 6.99 & 6.90 & \textbf{6.77} \\
        C4 PPL ($\downarrow$) & 9.56 & 9.24 & 9.00 & 8.91 & \textbf{8.71} \\
        Six zero-shot avg ($\uparrow$) & 60.32 & 61.35 & 61.35 & \textbf{62.30} & 62.22 \\
      \bottomrule
    \end{tabular}
\end{table}

%% file: tex/conclusion.tex
\section{Conclusion}
    In this paper, we introduce a novel adaptive framework that selects the optimal transformation on a layer-wise basis. Specifically, we present a differentiable search that formulates the problem as an optimization task to automatically find the best transformation for each layer, and then propose an efficient heuristic method. The heuristic leverages the connection between weight distribution characteristics and the optimal transformation type, achieving performance comparable to the differentiable search with significantly less computational overhead. This makes our approach practical for even the largest models. Our comprehensive experiments on the LLaMA family of models demonstrate the superiority of our adaptive approach. Compared to state-of-the-art methods like FlatQuant, our technique achieves a notable improvement, reducing perplexity by up to 4.58 points on the LLaMA-3-8B model under an aggressive W3A3KV2 quantization setting. These results validate the importance of layer-specific transformations for effective LLM quantization and offer a scalable solution for practical deployment.

%% file: reference.bib
@article{
  Wei_2022,
title={Emergent Abilities of Large Language Models},
author={Jason Wei and Yi Tay and Rishi Bommasani and Colin Raffel and Barret Zoph and Sebastian Borgeaud and Dani Yogatama and Maarten Bosma and Denny Zhou and Donald Metzler and Ed H. Chi and Tatsunori Hashimoto and Oriol Vinyals and Percy Liang and Jeff Dean and William Fedus},
journal={TMLR},
year={2022},
}

@article{frantar2022gptq,
  title={Gptq: Accurate post-training quantization for generative pre-trained transformers},
  author={Frantar, Elias and Ashkboos, Saleh and Hoefler, Torsten and Alistarh, Dan},
  journal={ICLR},
  year={2023}
}

@article{lin2023awq,
  title   = {AWQ: Activation-aware Weight Quantization for LLM Compression and Acceleration},
  author  = {Lin, Ji and Tang, Jiaming and Tang, Haotian and Yang, Shang and Dang, Xingyu and Han, Song},
  journal={Proceedings of machine learning and systems},
  year    = {2024}
}

@inproceedings{xiao2023smoothquant,
  title={Smoothquant: Accurate and efficient post-training quantization for large language models},
  author={Xiao, Guangxuan and Lin, Ji and Seznec, Mickael and Wu, Hao and Demouth, Julien and Han, Song},
  booktitle={ICML},
  year={2023}
}

@inproceedings{wei2023outlier,
  title   = {Outlier Suppression+: Accurate quantization of large language models by equivalent and optimal shifting and scaling},
  author  = {Wei, Xiuying and Zhang, Yunchen and Li, Yuhang and Zhang, Xiangguo and Gong, Ruihao and Guo, Jinyang and Liu, Xianglong},
  booktitle = {EMNLP},
  year    = {2023}
}

@inproceedings{
Hendrycks_2020,
title={Measuring Massive Multitask Language Understanding},
author={Dan Hendrycks and Collin Burns and Steven Basart and Andy Zou and Mantas Mazeika and Dawn Song and Jacob Steinhardt},
booktitle={ICLR},
year={2021},
}

@article{zhang2022opt,
  title={Opt: Open pre-trained transformer language models},
  author={Zhang, Susan and Roller, Stephen and Goyal, Naman and Artetxe, Mikel and Chen, Moya and Chen, Shuohui and Dewan, Christopher and Diab, Mona and Li, Xian and Lin, Xi Victoria and others},
  journal={arXiv},
  year={2022}
}

@article{Dettmers,
 title={Gpt3. int8 (): 8-bit matrix multiplication for transformers at scale},
 author={Dettmers, Tim and Lewis, Mike and Belkada, Younes and Zettlemoyer, Luke},
 journal={NeurIPS},
 year={2022}
}

@article{yao2022zeroquant,
  title={Zeroquant: Efficient and affordable post-training quantization for large-scale transformers},
  author={Yao, Zhewei and Yazdani Aminabadi, Reza and Zhang, Minjia and Wu, Xiaoxia and Li, Conglong and He, Yuxiong},
  journal={NeurIPS},
  year={2022}
}

@article{raffel2020exploring,
  title={Exploring the limits of transfer learning with a unified text-to-text transformer},
  author={Raffel, Colin and Shazeer, Noam and Roberts, Adam and Lee, Katherine and Narang, Sharan and Matena, Michael and Zhou, Yanqi and Li, Wei and Liu, Peter J},
  journal={JMLR},
  year={2020}
}

@inproceedings{
merity2016pointer,
title={Pointer Sentinel Mixture Models},
author={Stephen Merity and Caiming Xiong and James Bradbury and Richard Socher},
booktitle={ICLR},
year={2017}
}

@article{paperno2016lambada,
  title={The LAMBADA dataset: Word prediction requiring a broad discourse context},
  author={Paperno, Denis and Kruszewski, Germ{\'a}n and Lazaridou, Angeliki and Pham, Quan Ngoc and Bernardi, Raffaella and Pezzelle, Sandro and Baroni, Marco and Boleda, Gemma and Fern{\'a}ndez, Raquel},
  journal={arXiv preprint arXiv:1606.06031},
  year={2016}
}

@article{shao2023omniquant,
  title={OmniQuant: Omnidirectionally Calibrated Quantization for Large Language Models},
  author={Shao, Wenqi and Chen, Mengzhao and Zhang, Zhaoyang and Xu, Peng and Zhao, Lirui and Li, Zhiqian and Zhang, Kaipeng and Gao, Peng and Qiao, Yu and Luo, Ping},
  journal={ICLR},
  year={2024}
}

@article{cui2024cherry,
  title={Cherry on Top: Parameter Heterogeneity and Quantization in Large Language Models},
  author={Cui, Wanyun and Wang, Qianle},
  journal={NeurIPS},
  year={2024}
}

@inproceedings{kim2023squeezellm,
  title   = {Squeezellm: Dense-and-sparse quantization},
  author  = {Kim, Sehoon and Hooper, Coleman and Gholami, Amir and Dong, Zhen and Li, Xiuyu and Shen, Sheng and Mahoney, Michael W and Keutzer, Kurt},
  booktitle = {ICML},
  year    = {2024}
}

@article{touvron2023llama,
  title={Llama 2: Open foundation and fine-tuned chat models},
  author={Touvron, Hugo and Martin, Louis and Stone, Kevin and Albert, Peter and Almahairi, Amjad and Babaei, Yasmine and Bashlykov, Nikolay and Batra, Soumya and Bhargava, Prajjwal and Bhosale, Shruti and others},
  journal={arXiv},
  year={2023}
}

@inproceedings{
liu2025spinquant,
title={SpinQuant: {LLM} Quantization with Learned Rotations},
author={Zechun Liu and Changsheng Zhao and Igor Fedorov and Bilge Soran and Dhruv Choudhary and Raghuraman Krishnamoorthi and Vikas Chandra and Yuandong Tian and Tijmen Blankevoort},
booktitle={ICLR},
year={2025},
}

@article{arc,
  title={Think you have solved question answering? try arc, the ai2 reasoning challenge},
  author={Clark, Peter and Cowhey, Isaac and Etzioni, Oren and Khot, Tushar and Sabharwal, Ashish and Schoenick, Carissa and Tafjord, Oyvind},
  journal={arXiv},
  year={2018}
}

@inproceedings{piqa,
  title={Piqa: Reasoning about physical commonsense in natural language},
  author={Bisk, Yonatan and Zellers, Rowan and Gao, Jianfeng and Choi, Yejin and others},
  booktitle={AAAI},
  year={2020}
}

@inproceedings{hellaswag,
    title = "{H}ella{S}wag: Can a Machine Really Finish Your Sentence?",
    author = "Zellers, Rowan  and
      Holtzman, Ari  and
      Bisk, Yonatan  and
      Farhadi, Ali  and
      Choi, Yejin",
    booktitle = "ACL",
    year = "2019"
}

@article{sakaguchi2021winogrande,
  title={Winogrande: An adversarial winograd schema challenge at scale},
  author={Sakaguchi, Keisuke and Bras, Ronan Le and Bhagavatula, Chandra and Choi, Yejin},
  journal={Communications of the ACM},
  volume={64},
  number={9},
  pages={99--106},
  year={2021},
}

@article{ashkboos2024quarot,
  title={Quarot: Outlier-free 4-bit inference in rotated llms},
  author={Ashkboos, Saleh and Mohtashami, Amirkeivan and Croci, Maximilian and Li, Bo and Cameron, Pashmina and Jaggi, Martin and Alistarh, Dan and Hoefler, Torsten and Hensman, James},
  journal={NeurIPS},
  year={2024}
}

@article{tseng2024quip,
  title={Quip\#: Even better llm quantization with hadamard incoherence and lattice codebooks},
  author={Tseng, Albert and Chee, Jerry and Sun, Qingyao and Kuleshov, Volodymyr and De Sa, Christopher},
  journal={ICML},
  year={2024}
}

@inproceedings{
chee2023quip,
title={Qu{IP}: 2-Bit Quantization of Large Language Models With Guarantees},
author={Jerry Chee and Yaohui Cai and Volodymyr Kuleshov and Christopher De Sa},
booktitle={NeurIPS},
year={2023}
}

@inproceedings{lee2024owq,
  title={Owq: Outlier-aware weight quantization for efficient fine-tuning and inference of large language models},
  author={Lee, Changhun and Jin, Jungyu and Kim, Taesu and Kim, Hyungjun and Park, Eunhyeok},
  booktitle={AAAI},
  year={2024}
}

@inproceedings{hu2025ostquant,
  title={Ostquant: Refining large language model quantization with orthogonal and scaling transformations for better distribution fitting},
  author={Hu, Xing and Cheng, Yuan and Yang, Dawei and Xu, Zukang and Yuan, Zhihang and Yu, Jiangyong and Xu, Chen and Jiang, Zhe and Zhou, Sifan},
  booktitle={ICLR},
  year={2025}
}

@inproceedings{sun2024flatquant,
  title={Flatquant: Flatness matters for llm quantization},
  author={Sun, Yuxuan and Liu, Ruikang and Bai, Haoli and Bao, Han and Zhao, Kang and Li, Yuening and Hu, Jiaxin and Yu, Xianzhi and Hou, Lu and Yuan, Chun and others},
  booktitle={ICML},
  year={2025}
}

@book{iglewicz1993volume,
  title={Volume 16: how to detect and handle outliers},
  author={Iglewicz, Boris and Hoaglin, David C},
  year={1993},
  publisher={Quality Press}
}

@inproceedings{AffineQuant,
  author={Yuexiao Ma and Huixia Li and Xiawu Zheng and Feng Ling and Xuefeng Xiao and Rui Wang and Shilei Wen and Fei Chao and Rongrong Ji},
  title={AffineQuant: Affine Transformation Quantization for Large Language Models},
  year={2024},
  booktitle={ICLR},
}

@inproceedings{
an2025systematic,
title={Systematic Outliers in Large Language Models},
author={Yongqi An and Xu Zhao and Tao Yu and Ming Tang and Jinqiao Wang},
booktitle={ICLR},
year={2025},
}

@inproceedings{quik,
    title = "{QUIK}: Towards End-to-end 4-Bit Inference on Generative Large Language Models",
    author = {Ashkboos, Saleh and
      Markov, Ilia  and
      Frantar, Elias  and
      Zhong, Tingxuan  and
      Wang, Xincheng  and
      Ren, Jie  and
      Hoefler, Torsten  and
      Alistarh, Dan},
    booktitle ={EMNLP},
    month = {11},
    year = {2024},
    publisher = "Association for Computational Linguistics",
    doi = "10.18653/v1/2024.emnlp-main.197",
    pages = "3355--3371"
}

@article{decarlo1997meaning,
  title={On the meaning and use of kurtosis.},
  author={DeCarlo, Lawrence T},
  journal={Psychological methods},
  year={1997},
  publisher={American Psychological Association}
}

@inproceedings{
becigneul2018riemannian,
title={Riemannian Adaptive Optimization Methods},
author={Gary Becigneul and Octavian-Eugen Ganea},
booktitle={ICLR},
year={2019},
}
